\documentclass{acm_proc_article-sp}
\usepackage{fonttable}
\usepackage{url}
\usepackage{mathptmx}
\usepackage{color}
\usepackage{subfig}

\newlength{\figurewidthA} 
\newlength{\figurewidthB} 
\newlength{\figurewidthC} 
\newlength{\figurewidthE} 
\newlength{\figurewidthF} 
\newlength{\figurewidthG} 
\newlength{\figurewidthH} 
\newlength{\figurewidthI} 
\allowdisplaybreaks
\begin{document}

\twocolumn[%
  \begin{center}
    {\LARGE\textbf{\textsf{Object Localization and Size Estimation from RGB-D Images}}}\\
      \medskip
      {\large{S. SrirangamSridharan$^{*}$ \hspace{0.2in} O. Ulutan$^{+}$ \hspace{0.2in} S. N. T. Priyo${^{*}}$ \hspace{0.2in} S. Rallapalli$^{*}$ \hspace{0.2in} M. Srivatsa$^{*}$}}\\
      \medskip
      {\normalsize{IBM T. J. Watson Research Centre$^{*}$ \hspace{0.3in} UC Santa Barbara$^{+}$}}
      \end{center}
  ]

\begin{abstract}
Depth sensing cameras (e.g., Kinect sensor, Tango phone) can acquire 
color and depth images that are registered to a common viewpoint. 
This opens the possibility of developing algorithms that exploit the 
advantages of both sensing modalities. Traditionally, cues from color 
images have been used for object localization (e.g., YOLO). However, 
the addition of a depth image can be further used to segment images 
that might otherwise have identical color information. Further, the 
depth image can be used for object size (height/width) estimation (in real-world 
measurements units, such as meters) as opposed to image based 
segmentation that would only support drawing bounding boxes around 
objects of interest. In this paper, we first collect color camera information
along with depth information using a custom Android application on Tango 
Phab2 phone. Second, we perform timing and spatial alignment between the two
data sources. Finally, we evaluate several ways of measuring the height of 
the object of interest within the captured images under a variety of settings.
\end{abstract}

\section{Introduction}
\label{sec:intro}

Object detection involves recognizing an object present in an image (classification) 
as well as localization in order to compute a bounding box around it (localization). It is 
a well studied problem in computer vision~\cite{yolo,rcnn,fast_rcnn,faster_rcnn}. These techniques
typically leverage the color camera information alone. 
In many environments, Convolutional
Neural Networks (CNNs) based techniques have exhibited very high accuracy
of object detection. Yet, these techniques are vulnerable to errors due to
the quality of color camera images, variations in lighting conditions,  
and the absence of color based differentiation
in order to detect the object. 

On the other hand, depth images are becoming available these days due to the use of
Infrared (IR) sensor in devices like the Phab2 Tango phone as well as the Kinect. IR
sensors are documented to exhibit errors of $\sim$4cms at a range of 5m~\cite{ir_accuracy}. 
IR sensors are very energy efficient and can be mounted on mobile devices. On the other hand,
for higher accuracies one could also leverage the more expensive LIDAR sensors that are commonly utilized 
by autonomous driving systems. LIDAR sensors are known to have have lower errors 
$\sim$2cms at 100m range~\cite{lidar_accuracy}. Depth images can enhance the information from RGB cameras by providing the shape information
t

Further, object detection based on color images alone can provide the relative
size of the image in terms of pixels, however, this has no relation to real world
dimensions of the image. A toy car near the color camera may appear as big as a 
real car that is far away from the camera. Depth images also provide us with an 
opportunity to map the detected pixels to the real world size -- in terms of 
a measurement unit like meters. 

Estimating the real world dimensions of an object can be useful for many applications.
For instance, real word dimensions in turn can be used to improve the accuracy of 
object classification. For example, an object classified as a 6 feet tall dog, could
in fact be a horse -- incorrectly classified as a dog. Other applications of size estimation
include livestock health monitoring~\cite{livestock_health}, augmented reality and home furnishing. 

In this paper, we first collect multi-modal data using a custom Android application written for Tango
Phab2 phones. Second, we perform data pre-processing to perform spatio-temporal alignment of depth sensor (IR sensor)
information with the color camera information. Spatial alignment is required due to the difference
in the position of the color camera and the depth camera on the mobile device. Temporal 
alignment is required due to the time-lapse between the collection of the color camera data and the depth camera data. For e.g.,
on the Phab2 phone, depth camera sampling frequency is about 5 Hz, whereas the color camera data is collected
at about 30Hz. Thus we need to account for the difference in the position of the device at the different instances
that depth information and RGB information is collected.
After this alignment, we have depth information for some of 
the pixels of the color camera images. To obtain depth information for all the pixels we perform
bi-linear interpolation and use KD-tree for a fast search of nearest pixel with depth information. 
Third, we implement several techniques to compute the dimensions of the object of interest in meters 
and show that it works in a variety of environments both indoors and outdoors.

We summarize the contributions of this paper as follows:
\begin{itemize}
\item We collect multi-model data using a custom Android application and perform data alignment between
color camera and IR sensor data. 
\item To obtain depth information at all the pixels, the paper performs fast bi-linear clustering based on a spatial index in
the form of a KD tree.
\item The paper implements and evaluates several algorithms for accurately measuring object dimensions using this multi-model data.
\end{itemize}
 
In the rest of the paper, we explain our approach to fuse color camera information
with the depth images, followed by object segmentation and size-estimation. We 
overview the related work in Section \ref{sec:related}. We describe our approach and present in results 
in Section \ref{sec:approach}.  We conclude in Section \ref{sec:conclusion}.

\section{Related Work}
\label{sec:related}


Our work is related to the research in two specific areas: (i) object
detection and segmentation using color images, and (ii) fusing color
camera images with depth images for accurate object localization.
We overview some of the representative works in these areas below. 

\textbf{Object detection and segmentation using color images:}
There is a lot of work in the area of object detection and segmentation
in computer vision literature~\cite{yolo,rcnn,fast_rcnn,faster_rcnn,fcn}. 
The algorithms have different accuracies and at the same time varying 
amount of compute requirements -- running time, memory, power. These provide 
methods to process and infer images, however they do not by themselves process 
depth images. We build upon on the YOLO pipeline since it can process frames in 
real time and is thus amenable for use in practical systems.

\textbf{RGB-D images for object localization:}
\cite{voting1, vote3deep} directly process 3D point cloud data and
propose optimizations to enable processing 3D data more efficiently,
using the fact that most space in 3D grid would be empty. Object
detection schemes like R-CNN~\cite{rcnn} use object proposals (in 2D)
and run classification on object proposals. Along similar lines 
\cite{3dproposal1, 3dproposal2, multiview, monocular3d} use 3D object proposals
to perform 3D localization of the objects. MV3D~\cite{multiview}
fuses different views (birds eye view and frontal) of the point 
cloud information for accurate object localization. While all of 
these techniques are promising, they are computationally intensive 
since they typically use a multi-stage pipeline involving proposal
generation followed by classification and localization. \cite{2dto3d}
estimates 3D bounding box from 2D bounding box. \cite{rgbd_hha} 
performs late fusion of RGB with HHA features (obtained from point-cloud
data) for accurate object detection. While this is accurate, this
method requires more parameters due to the late fusion of extracted features.

\section{Object localization and Size Estimation}
\label{sec:approach}

We descibe three approaches to object localization and size estimation 
below: (i) one that uses depth camera information alone, 
(ii) that requires no re-training of existing RGB based object
detectors, can leverage off-the-shelf detectors and perform
size estimation using depth camera information towards the end 
of the estimation pipeline and (iii) that performs early fusion of RGB and depth
camera information and requires re-training of object detectors to leverage
the depth camera information. While the third scheme is more accurate, the first
two schemes are easy to deploy and more practial. 
We show the results from scheme (i) and (ii) while scheme (iii) is work in progress.

For all of the above schemes, we need to perform the alignment described below.

\subsection{Data Alignment}
This step of the pipeline involves aligning the point cloud data obtained
from the depth camera with the color camera data.
Point cloud data contains 
the real world co-ordinates of the pixels in meters --
i.e., height above ground, depth as well as the horizontal displacement from 
the center of the LIDAR/IR sensor. However, point-cloud data needs to be 
spatially and temporally transformed to RGB space in order to find matching 
point-cloud data for every RGB pixel. Point-cloud information maybe collected
at a different rate compared to the RGB camera, for example, Lenovo Tango Phab 2 
phone collects point cloud data at 5fps where as RGB data at 30 fps. Moreover,
there is also a spatial shift between the RGB camera and IR depth sensor that needs 
to be accounted for. After performing the transformation, every pixel will have 
R, G, B as well as real world X, Y, Z (in meters) information. 


\textbf{Mapping point cloud data to RGB space:} As a first step to perform this mapping, 
we calculate the transform (translation and
rotation) between the color camera at the time the user clicked and the depth camera 
at the time the depth cloud was acquired. This accounts for difference in time and 
frame of reference of the two sensors.

After the above step, we have the point cloud data in the color camera co-ordinate frame of
reference. Next, we project the point cloud data to the camera plane as follows. 
Given a 3D point $(X, Y, Z)$ in camera coordinates, 
the corresponding pixel coordinates $(x, y)$ are~\cite{tango_transform}: 
\begin{equation}
x = \frac{X}{Z} * fx * \frac{rd}{ru} + cx
\end{equation}

\begin{equation}
y = \frac{Y}{Z} * fy * \frac{rd}{ru} + cy
\end{equation}

The normalized radial distance $ru$ is given by: 
$$ru = \sqrt\frac{X^2 + Y^2}{Z^2}$$.

The distorted radial distance $rd$ depends on 3 distortion coefficients 
$k1$, $k2$ and $k3$ exposed by the device: 
$$rd = ru + k1 \cdot ru^3 + k2 \cdot ru^5 + k3 \cdot ru^7$$ 
 
$fx$, $fy$ are the focal lengths of the camera in pixels along the x and y axis
respectively. $cx$, $cy$ are the principal point offsets in pixels along the 
x and y axis respectively.

\textbf{Computing XYZ information for every pixel:} Once the above mapping
is done, all point cloud points obtained are associated with some pixel
value $(x, y)$. However, we require every pixel on the image to be associated
with R, G, B as well as real word X, Y, Z, measurements. Point cloud information
obtained is typically sparse, for e.g., one data point per $10 \times 10$ pixel 
grid. To obtain X, Y, Z for every pixel, we follow one of the two approaches 
described below. Note that both of the below approaches, involve nearest
neighbor search to find the nearest pixel that has associated X, Y, Z information.
A linear scan on the pixels can be very expensive as the image resolution is 
$1080 \times 1920$ leading to over 2 million operations. To speed up the search,
we index the point cloud data in a $k$-d tree, there by speeding up the search
significantly ($\sim100x$).

(i) \textit{Nearest neighbor:} A straight-forward approach to assign point-cloud
information to a pixel $(x,y)$ is to search for the nearest pixel (say $(x_j,y_j)$) 
with point-cloud information and simply assign the $(X, Y, Z)$ corresponding to
$(x_j,y_j)$ to $(x,y)$. This is relatively fast but can be less accurate.

\begin{figure}[h]
\centering
\includegraphics[width=2.5in]{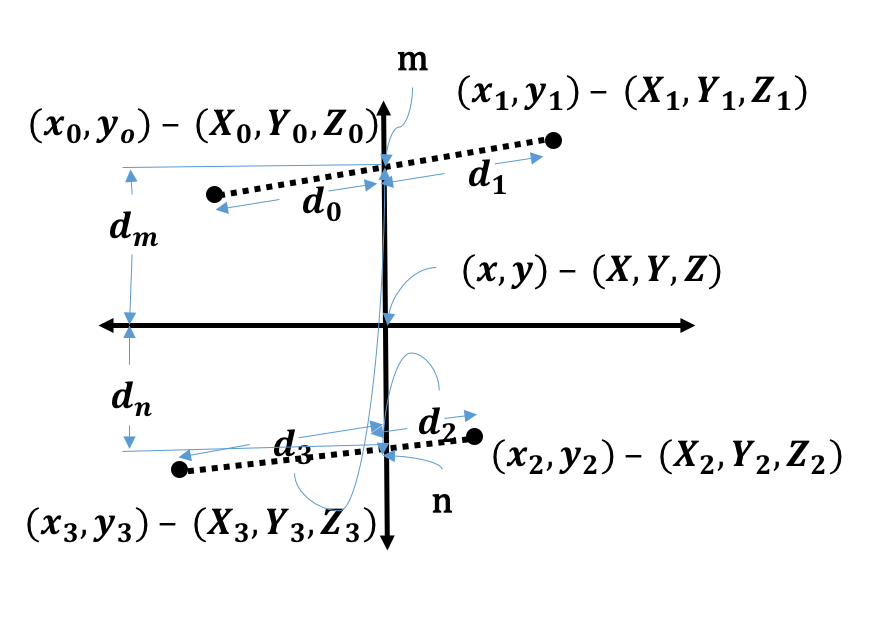}
\vspace{-0.2in}
\caption{Bi-linear interpolation.}
\label{fig:interp}
\end{figure}

(ii) \textit{Bi-linear interpolation:} A more accurate approach would be to find 
the four closest pixels with point-cloud information (i.e., closest pixel
in all four quadrants of a 2D space with origin as $(x,y)$) and iterpolate the $(X, Y, Z)$
values on the rectilinear grid. For instance see Figure~\ref{fig:interp}.
Convention we have been using is as follows: lower case, x, y correspond to pixel co-ordinates 
and upper case: X, Y, Z correspond to point cloud point i.e. real world co-ordinates. 
Suppose we want to find the point cloud point (X, Y, Z) corresponding to pixel (x, y): 
i.e., X, Y, Z --> unknown. Find closest points (using $k$-d tree for 100X speed up) 
in each of the 4 quadrants centered at (x, y) that have associated point cloud points. 
Then find m and n shown in the figure, using linear interpolation as shown below:
$$x_m = x_n = x$$
$$y_m = y_0 + (x - x_0) \cdot \frac{y_1 - y_0}{x_1 - x_0}$$
$$y_n = y_3 + (x - x_3) \cdot \frac{y_2 - y_0}{x_2 - x_3}$$

After finding pixels corresponding to m and n, euclidean distances $d_0, d_1, d_2, d_3$ are
found. For example, $d_0 = \sqrt{(x_m - x_0)^2 + (y_m - y_0)^2}$.
Then $X_m$ can be computed as $\frac{d_1 \cdot X_0 + d_0 \cdot X_1}{d_0+d_1}$
and similarly $Y_m, Z_m, X_n, Y_n$ and $Z_n$. Next, we compute $d_m$ and $d_n$ from
$(x,y)$, $(x_m,y_m)$ and $(x_n,y_n)$ and $X, Y, Z$ given as:
 $$X = \frac{d_m \cdot X_n + d_n \cdot X_m}{d_m+d_n}, Y = \frac{d_m \cdot Y_n + d_n \cdot Y_m}{d_m+d_n}, Z = \frac{d_m \cdot Z_n + d_n \cdot Z_m}{d_m+d_n} $$ 

After this step, every pixel $(x, y)$ has
a corresponding point cloud point $(X, Y, Z)$.
As an optimization, we can also preform bi-linear interpolation with edge detection, to interpolate
only if the points lie on the same surface as the pixel $(x,y)$ whose
point-cloud information is required. 

\begin{figure}[h]
\centering
\includegraphics[width=3in]{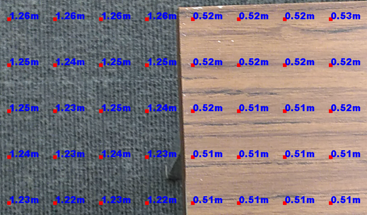}
\caption{After alignment of depth information with RGB information, we see the difference between
depth of table and depth of the floor from the camera -- in an image taken from above the desk.}
\label{fig:align}
\end{figure}

Figure \ref{fig:align} shows image taken from above a desk. The table is approximately 
0.5 meters from the camera whereas the floor is approximately 1.3 meters from the camera.

\subsection{Size Estimation}
Here we describe the different methods we use for object segmentation and size (object height) estimation.

\textbf{Scheme 1: Depth based clustering} \\
Our first scheme involves using depth
information alone to cluster the pixels. This enables us to obtain a tight boundary 
for an object. Specifically, we cluster the pixels of the image into 3 clusters: background, foreground,
and pixel with no depth information (typically filled in by 0s), using kmeans clustering from SciPy \cite{scipy_kmeans}.
Depth of the pixels are used for computing the distance measure in the kmeans algorithm. We pick the
second cluster as the foreground cluster. The object of interest is contained in the second cluster due
to the ordering of depth information (pixels with no depth information are at depth 0, followed by foreground pixels, followed by background pixels which are the farthest).
  
\begin{figure}[h]
\centering
\includegraphics[width=\figurewidthA]{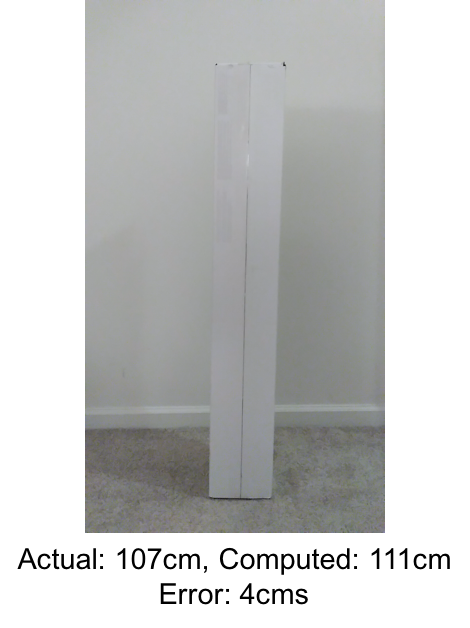}
\vspace{-0.2in}
\caption{Depth based clustering can segment out objects in cases where RGB based segmentation is challenging.}
\vspace{-0.1in}
\label{fig:white_box}
\end{figure}

In cases where there is only
a single object to be measured in an image, this approach is very useful. Interestingly,
this method can accurately estimate the dimensions of objects like the box shown in Figure~\ref{fig:white_box} \textit{where
color based differentiation is very challenging} due to the white object on the white background.
 
\begin{figure}[h!]
\centering
\subfloat{\includegraphics[width=\figurewidthG]{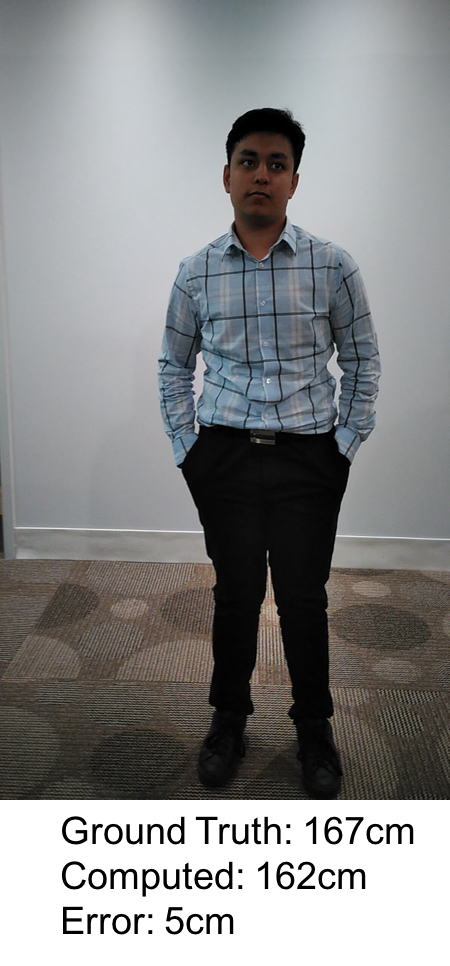}} \hspace{0.1in}
\subfloat{\includegraphics[width=\figurewidthG]{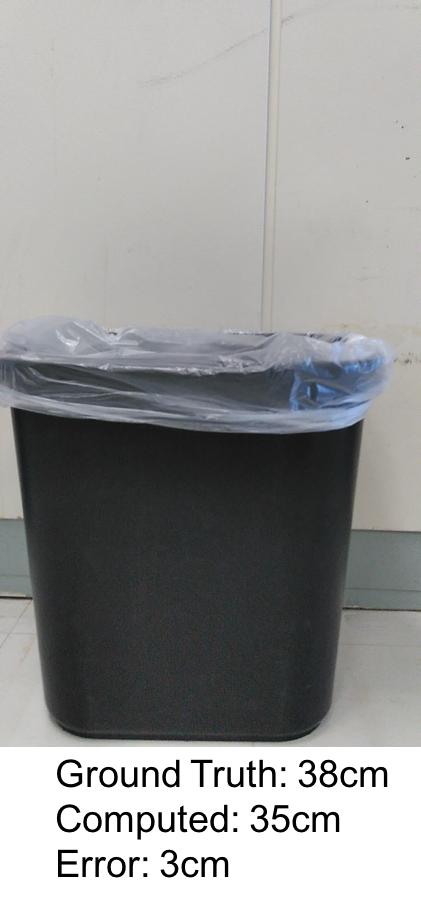}} \hspace{0.1in}
\subfloat{\includegraphics[width=\figurewidthG]{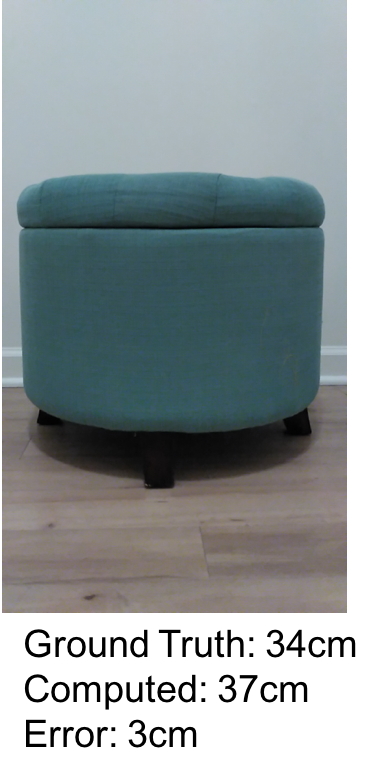}} \hspace{0.1in}
\subfloat{\includegraphics[width=\figurewidthG]{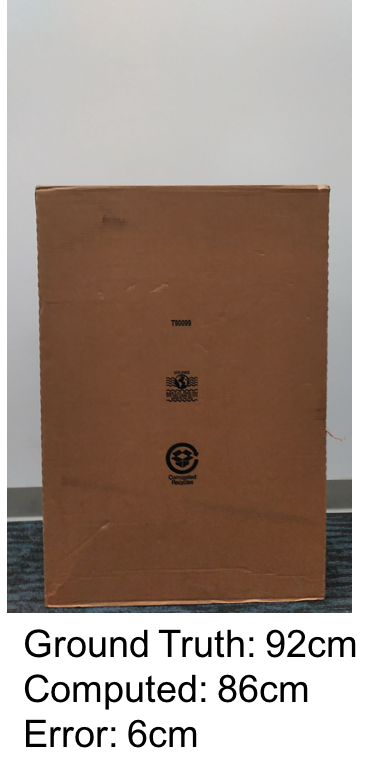}} \hspace{0.1in}
\vspace*{-0.1in}
\caption{Size estimation results based on depth based clustering alone.}
\vspace*{-0.1in}
\label{fig:scheme1_results}
\end{figure}
Figure~\ref{fig:scheme1_results} shows size estimation results on several other different subjects.
Note that the results are very accurate in cases where a single subject is present.
In cases where multiple objects are present in a cluttered scene, depth based clustering
alone is not sufficient and thus we use object detection along with depth based
clustering as described below.

\begin{figure}[h!]
\centering
\subfloat[Output after object detection.]{\includegraphics[width=\figurewidthH]{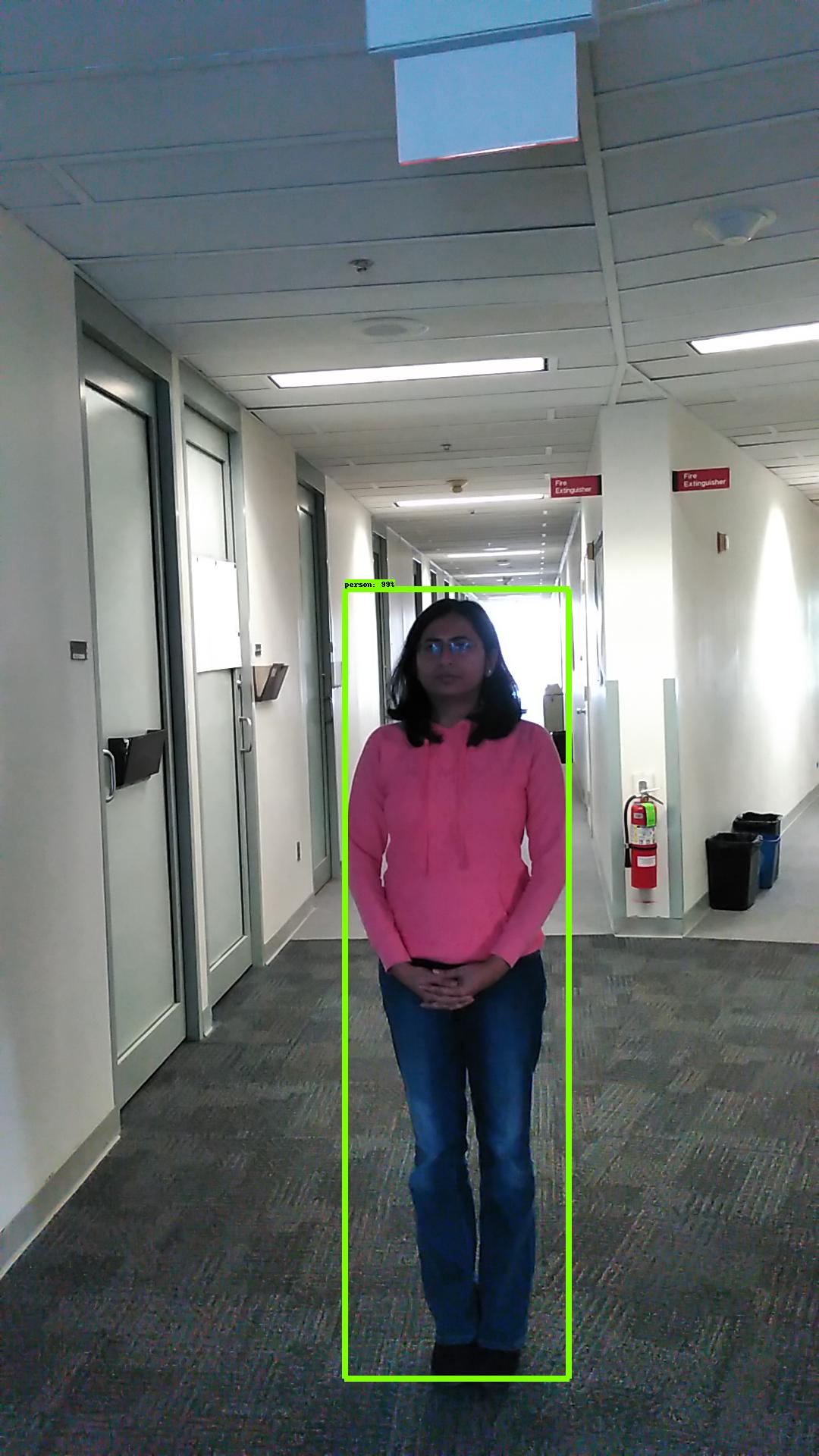}} \hspace{0.1in}
\subfloat[Output after depth based clustering.]{\includegraphics[width=\figurewidthA]{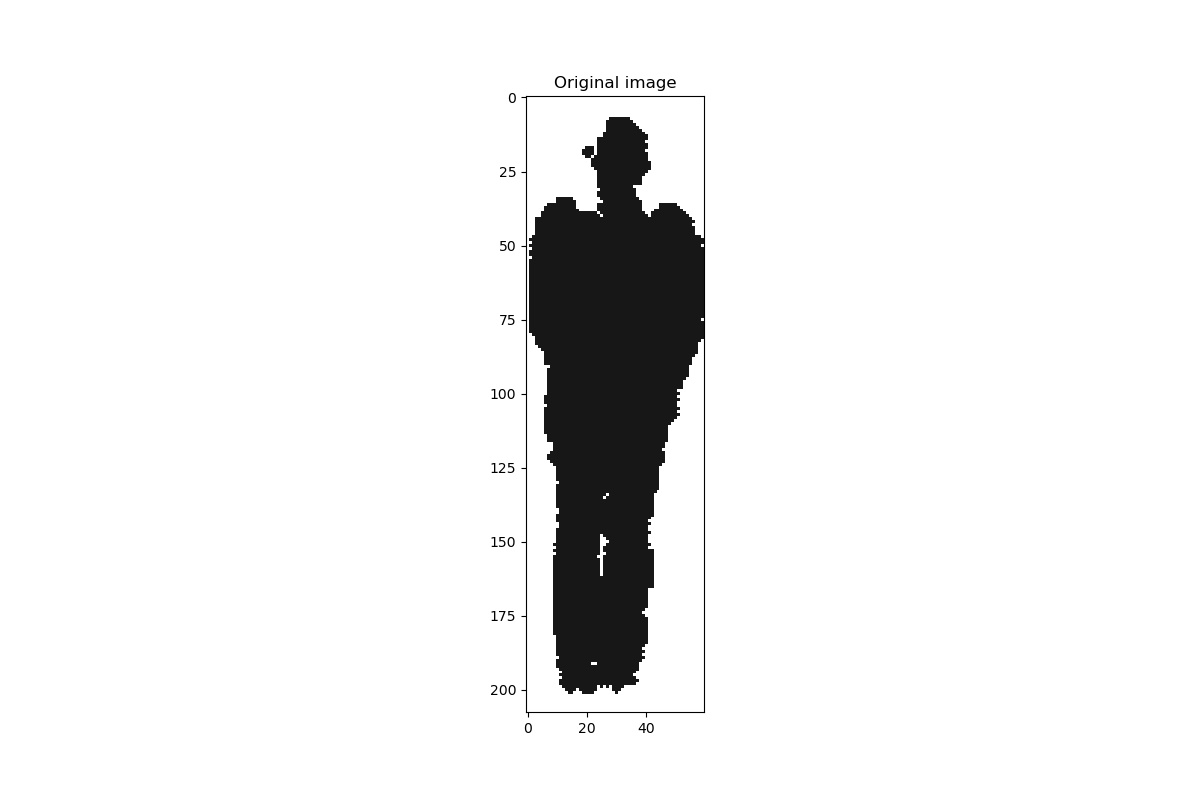}}
\vspace{-0.1in}
\caption{Intermediate results of object detection followed by depth based clustering.}
\vspace{-0.1in}
\label{fig:object_seg}
\end{figure}

\textbf{Scheme 2: RGB object detection + depth based clustering} \\
Since above explained depth based clustering alone is not sufficient when 
multiple objects are present in a scene, in this approach we first
leverage off-the-shelf pre-trained CNNs to detect the objects of interest within the scene.
We use off-the-shelf pre-trained CNNs so as to minimize the requirement of
labelled data for re-training. We use object detection 
from~\cite{tf_object_detection} to obtain bounding boxes around objects of interest.
This uses RGB information alone to obtain the bounding box. Different object detection algorithms 
can be used for RGB information based bounding box detection. The state-of-the-art schemes
are based on Deep Convolutional Neural Networks. 
Once the bounding box is obtained as shown in Figure~\ref{fig:object_seg} (a), to isolate the object more tightly and segment it out,
we cluster the pixels within the bounding box based on the depth information. An example 
clustered image is shown in Figure~\ref{fig:object_seg} (b). Once we have cleanly segmented
out the object, we are able to measure the dimensions based on the $(X,Y,Z)$ information
of the extreme points (pixels). 

\begin{figure}[h!]
\centering
\subfloat{\includegraphics[width=\figurewidthG]{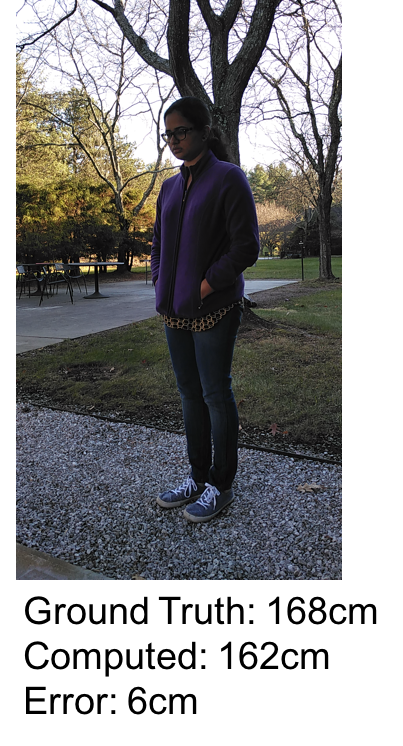}} \hspace{0.1in}
\subfloat{\includegraphics[width=\figurewidthG]{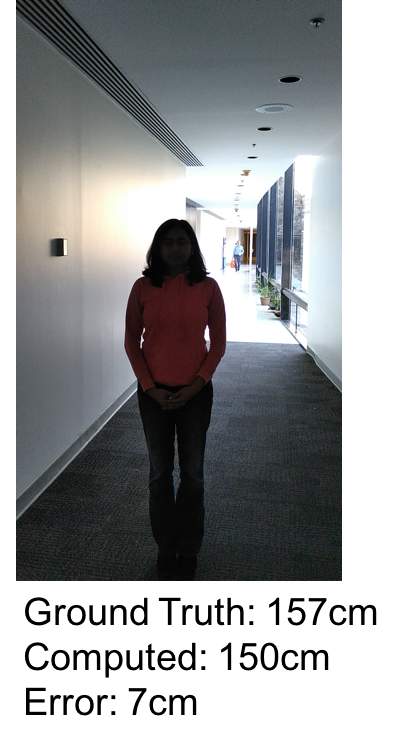}} \hspace{0.1in}
\subfloat{\includegraphics[width=\figurewidthG]{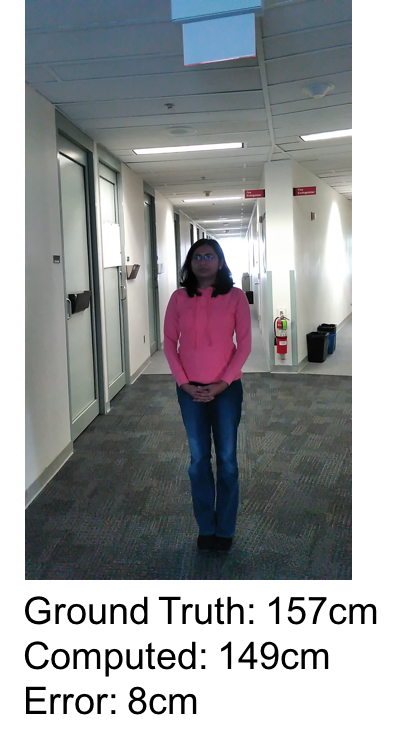}} \hspace{0.1in}
\subfloat{\includegraphics[width=\figurewidthG]{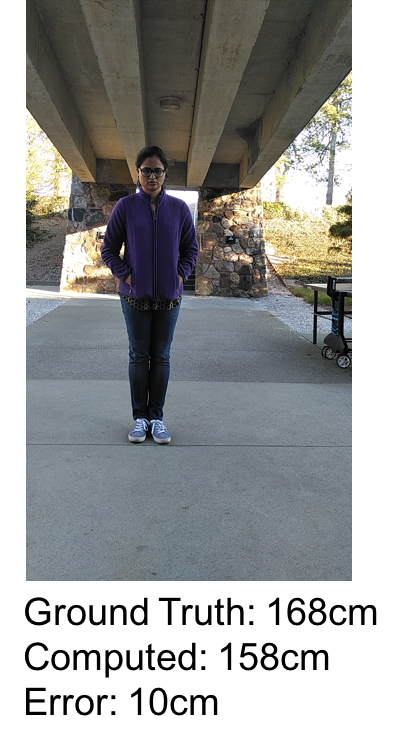}} \hspace{0.1in}
\vspace*{-0.1in}
\caption{Size estimation results based on object detection + depth based clustering. Experiments
were done both indoors and outdoors as seen in the images.}
\vspace*{-0.2in}
\label{fig:scheme2_results}
\end{figure}

Figure~\ref{fig:scheme2_results} shows size estimation results of object detection followed
by depth based clustering on humans. The scheme works accurately without requiring that the person
should stand against a clear background. 

\textbf{Scheme 3: Re-training YOLO for early fusion} \\
This is our on-going work. To further improve the accuracy of object localization, CNN pipelines can be
trained to fuse RGB features with depth based features. We choose YOLO pipeline
due to its efficiency and real-time nature. We obtain HHA features as described in
~\cite{rgbd_hha}. HHA channels are then fused with RGB to create a 6 channel image.
The early fusion minimizes the number of parameters in the CNN for high efficiency 
of computation. This improves the accuracy of localization because HHA features
can help where RGB is not distinguishing or clear enough for object detection.   
This approach combines the benefits of the previous approaches, i.e., (i) depth
information helps even in situations where color alone is not sufficient
and (ii) this approach can be used in cluttered scenes with multiple objects.  
Once the object is accurately detected, the depth based clustering and object size
estimation can be performed similar to above schemes.

\section{Discussion and Conclusion}
\label{sec:conclusion}
Object dimensions can be accurately estimated using color camera information
along with point cloud information. In order to do so accurately, we require
dense point cloud information of the object. In our experiments we found that 
point cloud information maybe missing in case of
black or metallic objects. Stereo imagery can help in cases where such data sparsity 
is a problem. Further, the accuracy of size estimation is 
higher in cases where background is not in close proximity of the object --
thus enabling accurate depth based segmentation.

\section{Acknowledgement}
Research was sponsored by the Army Research Laboratory and was
accomplished under Cooperative Agreement Number W911NF-09-2-0053
(the ARL Network Science CTA). The views and conclusions contained in this
document are those of the authors and should not be interpreted as representing
the official policies, either expressed or implied, of the Army Research
Laboratory or the U.S. Government. The U.S. Government is authorized
to reproduce and distribute reprints for Government purposes notwithstanding
any copyright notation here on.

{\small
\bibliographystyle{abbrv}
\bibliography{refs}  
}

\balancecolumns
\end{document}